\documentclass[10pt,twocolumn,letterpaper]{article}

\usepackage{cvpr}              %
\usepackage{amsmath}
\usepackage{soul}
\usepackage[accsupp]{axessibility}
\usepackage{algorithm}
\usepackage{algorithmic}
\usepackage[algo2e]{algorithm2e}
\SetKwComment{Comment}{$\triangleright$\ }{}
\usepackage{diagbox}

\newcommand{\halfcheckmark}{%
  \textcolor{black}{\ding{51}}{\small\textcolor{black}{\kern-0.7em\ding{55}}}
}

\newcommand{\newcheckmark}{%
  \textcolor{black}{\ding{51}}
}

\usepackage{pifont}

\definecolor{cvprblue}{rgb}{0.21,0.49,0.74}
\usepackage[pagebackref,breaklinks,colorlinks,allcolors=cvprblue]{hyperref}

\newcommand{\tb}[1]{\textcolor{cvprblue}{#1}}

\def\paperID{10925} %
\def\confName{CVPR}
\def\confYear{2025}

\title{Analyzing the Synthetic-to-Real Domain Gap in 3D Hand Pose Estimation}

\author{Zhuoran Zhao\textsuperscript{1}\thanks{Done while the first author was a research intern at NUS@CVML}
\hspace{3mm}
Linlin Yang\textsuperscript{2}\thanks{Corresponding author} \hspace{3mm}
Pengzhan Sun\textsuperscript{3} \hspace{3mm}
Pan Hui\textsuperscript{1}\hspace{3mm}
Angela Yao\textsuperscript{3} \hspace{3mm}\\%
\textsuperscript{1}The Hong Kong University of Science and Technology (Guangzhou), China \hspace{3mm} \\
\textsuperscript{2}Communication University of China, China \hspace{3mm} 
\textsuperscript{3}National University of Singapore, Singapore \\
\tt\small zzhao074@connect.hkust-gz.edu.cn,
\tt\small lyang@cuc.edu.cn,
\tt\small panhui@ust.hk, \\
\tt\small \{pengzhan, ayao\}@comp.nus.edu.sg
}

\begin{document}
\maketitle
\begin{abstract}
\indent Recent synthetic 3D human datasets for the face, body, and hands have pushed the limits on photorealism. Face recognition and body pose estimation have achieved state-of-the-art performance using synthetic training data alone, but for the hand, there is still a large synthetic-to-real gap. This paper presents the first systematic study of the synthetic-to-real gap of 3D hand pose estimation. We analyze the gap and identify key components such as the forearm, image frequency statistics, hand pose, and object occlusions. To facilitate our analysis, we propose a data synthesis pipeline to synthesize high-quality data. We demonstrate that synthetic hand data can achieve the same level of accuracy as real data when integrating our identified components, paving the path to use synthetic data alone for hand pose estimation. Code and data are available at: \href{https://github.com/delaprada/HandSynthesis.git}{https://github.com/delaprada/HandSynthesis.git}.
\end{abstract}    
\section{Introduction}
\label{sec:intro}

3D hand pose estimation from monocular RGB images plays an important role in computer vision, human-computer interaction, and AR/VR applications \cite{jiang2021skeleton, caggianese2020freehand, han2022umetrack}.  It is challenging since the hand has severe occlusion problems and depth ambiguity.  Most recent methods \cite{lin2021end-to-end, bib:CMR, moon2020i2l} rely heavily on 3D ground-truth labels for training, but obtaining these labels is expensive and time-consuming. Therefore, there is strong interest in using synthetic data, as it does not require any annotation effort.
In face recognition~\cite{wood2021fake, bae2023digiface} and 3D human pose estimation~\cite{black2023bedlam}, state-of-the-art is achievable with only synthetic training sets.  However, there is still a large synthetic-to-real gap for 3D hand pose estimation \cite{gao2022dart, li2023renderih, hasson2019learning, yang2021semihand}, despite recent efforts to diversify backgrounds, pose distributions, camera viewpoints, and hand textures for synthetic data.  Pre-training on synthetic and fine-tuning with real data, or mixed synthetic and real training, is still much better than training with only synthetic data ~\cite{gao2022dart, li2023renderih, moon2023reinterhand, chen2021mvhm, hasson2019learning}.
These observations beg the question: What contributes to the synthetic-to-real gap in 3D hand pose estimation?

\begin{figure}[tb]
  \centering
  \includegraphics[width=1\linewidth]{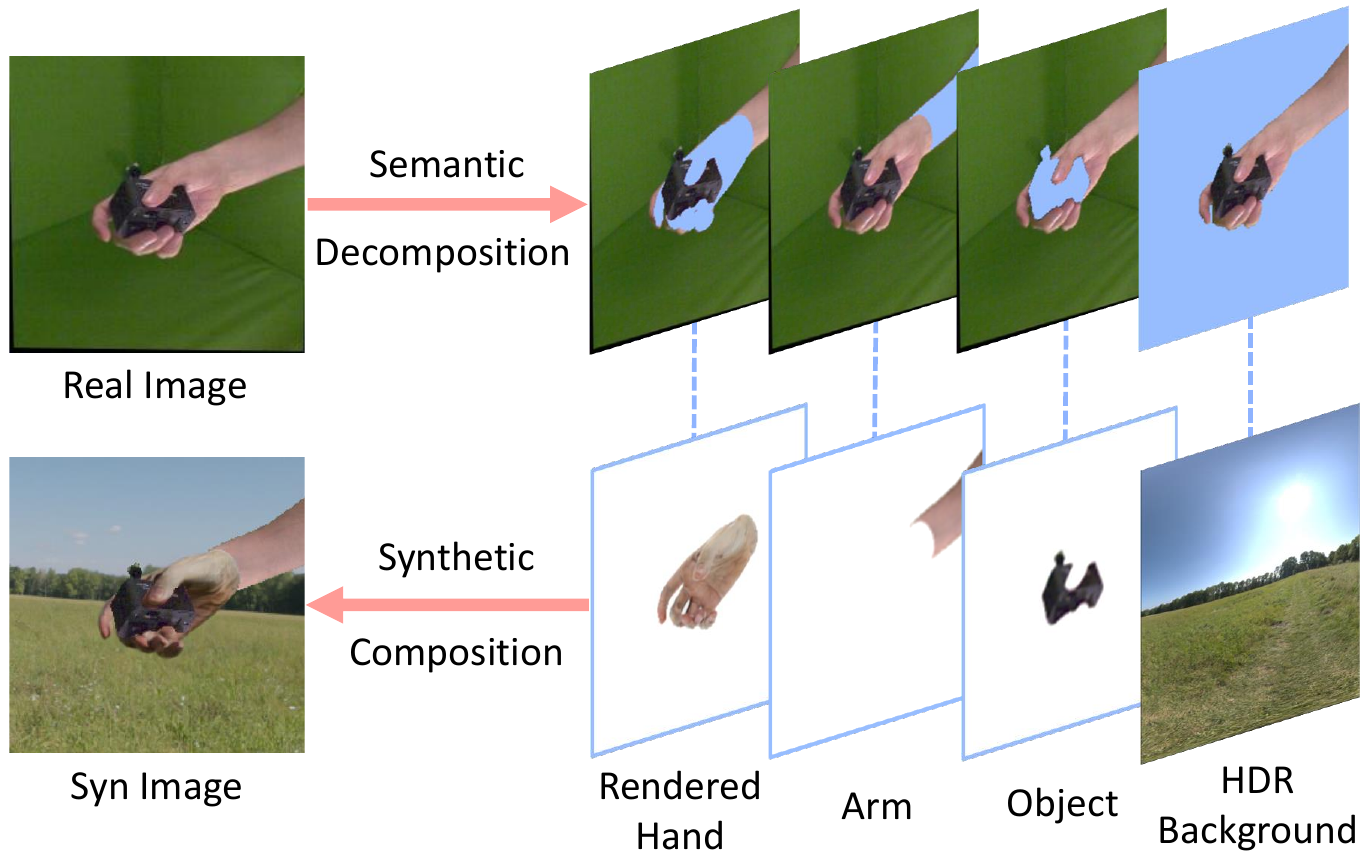}
  \caption{We present a systematic study of the synthetic-to-real gap for 3D hand pose estimation by decomposing the synthetic data to establish associations between hand image components and predictions.
  }
  \label{fig:header}
\end{figure}

Many synthetic hand datasets exist but are difficult to analyze since they differ from real hand datasets across multiple aspects (\eg, pose distribution, background, texture, \etc), which cannot be separated.  Therefore, we propose a data synthesis pipeline to facilitate our study.  The pipeline efficiently synthesizes high-quality data with diverse backgrounds, lighting, and hand textures.  With full control over the rendering pipeline, we can decompose synthetic data to analyze associations between different hand image components and predictions to enhance synthetic-to-real gap analysis (see Fig.~{\ref{fig:header}}).  Moreover, since our pipeline can synthesize controllable pair-wise data, it can benefit future research for further analysis and enhance model performance for downstream applications.

\begin{table*}[tb]
\centering
\caption{Representative synthetic hand datasets. HDR denotes high dynamic range scenes.  HDR Res. denotes HDR background resolution, with higher values being more realistic and less blurry.  For lighting, ``manual" denotes manually defined settings while ``dynamic" denotes diverse lighting conditions consistently applied across the foreground hand and background scene.  ``NS" means normal and specular maps for hand texturing.  ``GR" means the hand data generation and image rendering pipeline.}

\label{tab:syn_datasets}
\resizebox{1\textwidth}{!}{
\begin{tabular}{l|c|c|c|c|c|c|c|c|c|c}
\toprule
Dataset & Data Size & Background & HDR Type & HDR Res. & Lighting & Texture & NS & Template & Hand Type & GR \\
\midrule
RHD \cite{Zimmermann_2017_ICCV} & 44K & static  & - & -  & manual & limited & \ding{55}  & Mixamo & single hand & \ding{55} \\
ObMan \cite{hasson2019learning} & 154K & static  & - & - & manual & limited & \ding{55}  & MANO & hand object & \ding{51} \\
DARTset \cite{gao2022dart} & 800K & static  & - & -  & manual & diverse & \ding{55}  & DART & single hand & \ding{51} \\
Ego3D \cite{Lin_2021_WACV} & 55K & static  & - & -  & manual & limited & \ding{55}  & Mixamo & two hands & \ding{55} \\
Re:InterHand \cite{moon2023reinterhand} & 1.5M & HDR   & Indoor & 7K  & dynamic  & diverse & \ding{55} & MANO & two hands & \ding{55} \\
RenderIH \cite{li2023renderih} & 1M & HDR  & Indoor \& Outdoor & 1K  & dynamic & diverse & \ding{55}  & A-MANO & two hands & \ding{51} \\
\midrule
\textbf{Ours} & 460K & HDR  & Indoor \& Outdoor & 4K & dynamic & diverse & \ding{51}  & NIMBLE & 
single hand & \ding{51} \\

\bottomrule
\end{tabular}
}
\end{table*}

We analyze the synthetic-to-real gap from four aspects: appearance, hand pose, object occlusion, and hand skeleton topology, which are essential for hand pose estimation.
From the appearance perspective, we first find that the presence of the forearm is important for pose estimation models to locate the wrist, affecting overall pose accuracy. Secondly, amplitude spectrum augmentation~\cite{chattopadhyay2023pasta} is crucial for enhancing the model's robustness when the domain shifts from synthetic to real.  Thirdly, increasing the background and hand texture diversity is important in reducing the synthetic-to-real gap, but their influence saturates beyond a certain amount.
From the pose perspective, we observe that the pose difference between synthetic and real data affects the synthetic-to-real performance.  However, performance gains tend to plateau after a certain threshold as more real hand poses are used for data synthesis.  Notably, we find that using only a subset of real hand poses for synthesizing hand data enables the model trained on such data to achieve comparable performance to using all real hand poses.  This outcome demonstrates that the data synthesized by our pipeline effectively facilitates learning hand representations. 

Unlike the human face and body, the hand has greater self-occlusion and interacting object occlusions.  Yet, only a few synthetic hand datasets feature hand-object interactions.  
Incorporating object occlusions into synthetic training data improves model generalization on real-world test samples.  Additionally, we observe that pose estimation models can associate hand poses with specific objects.  Modeling specific object occlusions as a prior can improve performance when encountering  
similar objects. Finally, synthetic and real datasets often use different hand skeleton topologies~\cite{9262071, moon2024dataset}.  This discrepancy can cause inaccurate evaluation and label adaptation is necessary. 

By incorporating different hand image components, models trained with only our synthesized data can achieve state-of-the-art performance compared with real data.  Moreover, models trained on a mixture of real and synthetic data outperform those trained only on real data for in- and cross-domain generalization.  We summarize our contributions as follows:
\begin{enumerate}
    \item We present the first systematic study of the synthetic-to-real gap of 3D hand pose estimation.  Having full control over the rendering pipeline, we establish associations between hand-image components and predictions by decomposing synthetic data.
    \item We propose a high-quality synthetic data synthesis pipeline that supports our analysis and enables future applications to boost model performance. 
    \item We demonstrate that synthetic data can achieve the same level of accuracy as real data while paving the path to use synthetic data alone for hand pose estimation.  
\end{enumerate}

\section{Related Work}
\label{sec:related_work}

\textbf{Synthetic Hand Datasets.} Building real-world 3D hand datasets is a challenging and time-consuming task.  Recording often requires multi-camera systems or depth sensors~\cite{zimmermann2019freihand, YangCVPR2022OakInk, Moon_2020_ECCV_InterHand2.6M, hampali2020honnotate}.  After data collection, manual or semi-automated labeling is required for ground truth.  Synthetic data is a more time- and cost-effective alternative even if the images are not always photo-realistic; RHD~\cite{Zimmermann_2017_ICCV}, SynthHands~\cite{OccludedHands_ICCV2017} and ObMan~\cite{hasson2019learning} are three early examples of single hands, ego-centric hands, and hand grasping objects.  More recently, DARTset~\cite{gao2022dart} introduced a wrist-enhanced hand model and hands with different accessories.  Ego3D~\cite{Lin_2021_WACV}, Re:InterHand~\cite{moon2023reinterhand} and RenderIH~\cite{li2023renderih} feature two hands interacting, from both ego-centric~\cite{Lin_2021_WACV} and third-person views~\cite{moon2023reinterhand,li2023renderih}. The datasets differ in their skeleton topology, consideration of background, lighting, and hand texture for rendering (see Tab.~\ref{tab:syn_datasets}). The stand-alone efficacy of these synthetic datasets is unestablished, as most are used only for pre-training and then supplemented with either real or mixed data fine-tuning~\cite{gao2022dart, li2023renderih, moon2023reinterhand, chen2021mvhm, hasson2019learning}.  As we will show, these datasets still exhibit a significant synthetic-to-real gap; our work is the first systematic study to examine this gap for 3D hand pose estimation.

\noindent \textbf{Synthetic-to-Real 3D Hand Pose Estimation.} Several works have explored using synthetic data to enhance the generalization performance on real-world data for hand pose estimation.  In the depth domain, \cite{Wan_2019_CVPR, DeepHPS_3DV2018} estimated hand pose from synthetic depth.  For event data, Eventhands~\cite{rudnev2021eventhands} used synthetic event streams for training to capture hand poses.
Our work focuses on the RGB domain; one of the original works~\cite{Zimmermann_2017_ICCV} proposes a synthetic RGB dataset (RHD), which reports better generalization results than a limited stereo real hand dataset.  Other works learn from both labeled synthetic and unlabeled real-world data, using strategies such as pseudo-labeling and consistency training~\cite{yang2021semihand}, reconstruction-based approach pseudo-labels
~\cite{lin2024synthetic}, multi-modality supervisions~\cite{cai2018weakly, lin2023cross}, input and output representations alignment to build a unified framework for various domain adaptive pose estimation~\cite{kim2022unified}.

\begin{figure}[tb]
  \centering
  \includegraphics[width=1\linewidth]{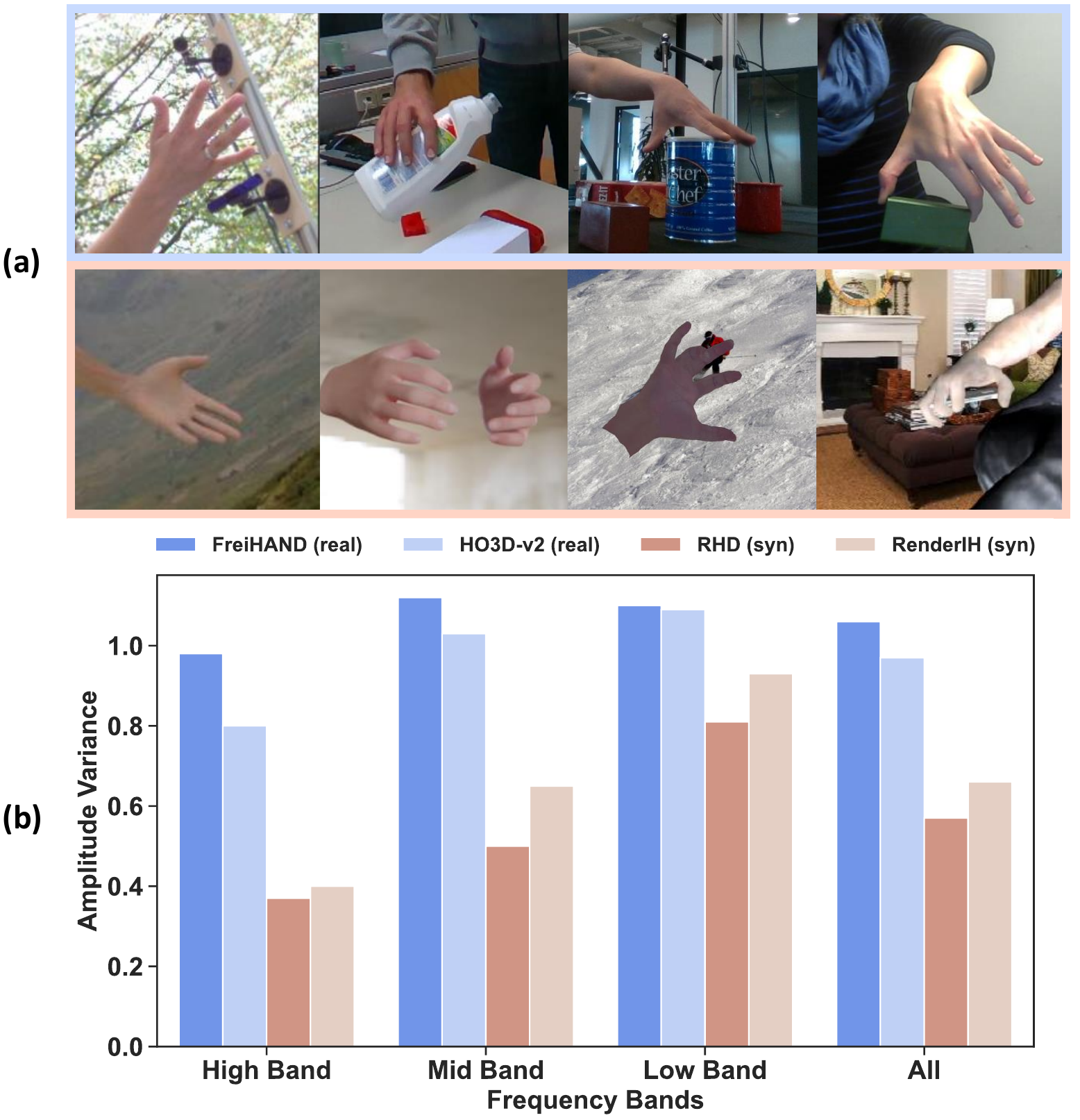}
  \caption{(a) Real images (blue) have richer hand texture information and complex environments than synthetic images (red).  Arms and interacting objects are often present in real images.  (b) Real datasets have higher variation in amplitude values than synthetic ones across all frequency bands.}
  \label{fig:real_syn}
\end{figure}

\section{Synthetic-to-Real Hand Analysis}

This work verifies the hand image components (Sec.~\ref{sec:component})
that contribute to the synthetic-to-real gap in hand pose estimation.  To that end, we develop a hand image synthesis pipeline (Sec.~\ref{sec:hand_gen}) for high-quality synthetic images that control the diversity of hand textures, backgrounds, and poses. We apply a compositional strategy (Sec.~\ref{sec:hand_decompose}) to show the impact of arms and interacting objects. Moreover, we show the connection between occlusion levels and the hand object prior (Sec.~\ref{sec:occ_prior}).

\subsection{Hand Image Component}\label{sec:component}

A typical hand image consists of the hand, the arm, interacting object(s), and the background. Comparing real hand images \cite{zimmermann2019freihand, hampali2020honnotate, chao2021dexycb, RealtimeHO_ECCV2016} with synthetic ones \cite{Zimmermann_2017_ICCV, li2023renderih, gao2022dart, hasson2019learning,yang2019disentangling}, we observe several differences across these components (see Fig.~\ref{fig:real_syn}(a)). We summarize the differences below. \\

\noindent\textbf{Appearance.} (1) The hands in real images come from multiple subjects and exhibit diverse hand textures; in synthetic images, the textures are limited and unrealistic. (2) Real images often feature complex and diverse environments, whereas synthetic images generally have simple and limited backgrounds.  
(3) Real images feature hands together with the arms, while synthetic images often only have a floating hand without the arm. 
(4) PASTA~\cite{chattopadhyay2023pasta} showed that synthetic images have less variance than real images in their high-frequency components for semantic segmentation.  We similarly analyze the variance of the amplitude in the frequency spectrum for real versus synthetic hand datasets (see Fig.~\ref{fig:real_syn}(b)). We find that synthetic hand datasets have less variance than real datasets across the entire frequency spectrum.\\

\noindent\textbf{Objects.} (1) Real images often feature hand-object interactions, leading to various occlusions. (2) Some objects may appear in both real train and test sets, while others are never seen in the synthetic datasets. Therefore, the synthetic datasets may lack the prior information on object occlusion that is present in the real datasets.\\

\noindent\textbf{Skeleton Topology.} The skeletons may differ for real and synthetic hand data. The skeleton topology is a systematic difference between human-annotated and synthetic labels created by parametric or commercial models. 

\subsection{Hand Image Synthesis}\label{sec:hand_gen}
\begin{figure}[tb]
  \centering
  \includegraphics[width=1\linewidth]{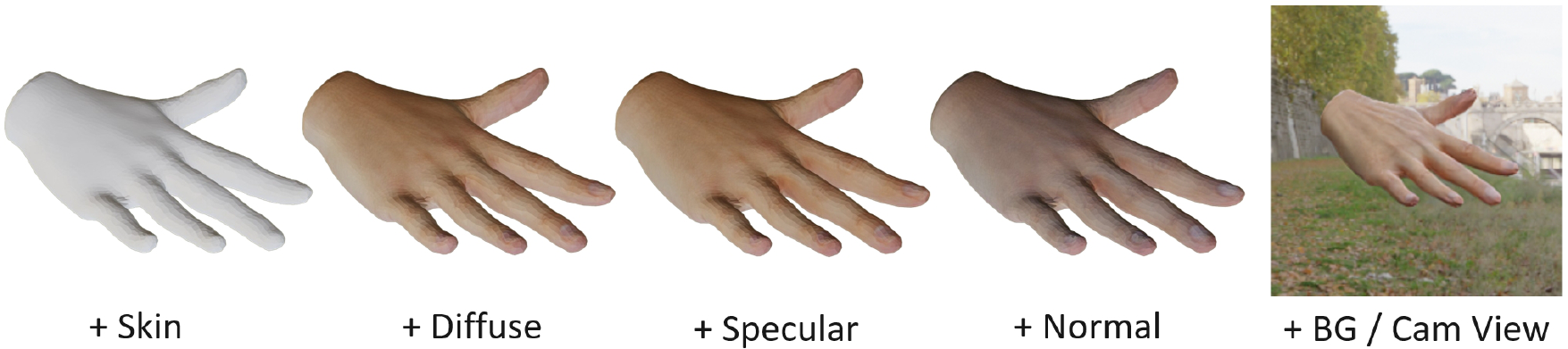}
  \caption{Image synthesis process.}
  \label{fig:syn}
\end{figure}

We propose a rendering process (see Fig.~\ref{fig:syn}) starting with a parametric hand model that models bones, muscles, skins, and textures. We then sample hand poses from real-world hand datasets. To push the boundary of realism, each hand is `dressed up' with high-resolution textures and dynamic backgrounds under varying lighting conditions.

\noindent\textbf{Hand Model.} We choose the NIMBLE~\cite{li2022nimble} hand model, which takes pose $\theta$, shape $\beta$ and texture $\alpha$ parameters as inputs and generates a realistic 3D hand:
\begin{equation}
    \mathcal{N}(\theta, \beta, \alpha) = \{\mathcal{G}(\theta, \beta), \mathcal{A}(\alpha)\}.
\end{equation}
\noindent Above, $\mathcal{G}(\theta, \beta)$ denotes the hand geometry, including skin, muscle, and bone meshes. The term $\mathcal{A}(\alpha)$ models the hand texture, including diffuse, specular, and normal maps. Most existing synthetic hand datasets~\cite{hasson2019learning, li2023renderih,gao2022dart} use the MANO hand model~\cite{romero2022embodied}. NIMBLE has a finer 3D mesh, with 5,990 vertices and 9,984 triangular faces, compared to the 778 vertices and 1,538 faces of MANO. NIMBLE and MANO have different topologies, including differences in both the number of joints (25 vs. 21) and their positions (See Supp.~Sec.~\tb{F}).

\noindent\textbf{Hand Pose.}
We keep our pose distribution consistent with the target real datasets (\ie, FreiHAND~\cite{zimmermann2019freihand} and Dex-YCB~\cite{chao2021dexycb}) to eliminate its influence during our analysis. 
We derive the hand pose by fitting the NIMBLE hand mesh to the MANO hand mesh. During the fitting process, we optimize the pose, shape, rotation, and translation parameters in two stages to avoid poor fitting results (See Supp.~Sec.~\tb{B}). Since our data synthesis pipeline is controllable and generalizable, we further augment our pose distribution with the fitting algorithm in three ways: (1) sampling poses from ObMan~\cite{hasson2019learning} to incorporate more hand-object interaction, (2) randomly synthesizing poses from NIMBLE space, and (3) sampling from VAE prior built with depth-based hand datasets~\cite{armagan_2020_ECCV}.

\begin{figure}[tb]
  \centering
  \includegraphics[width=1\linewidth]{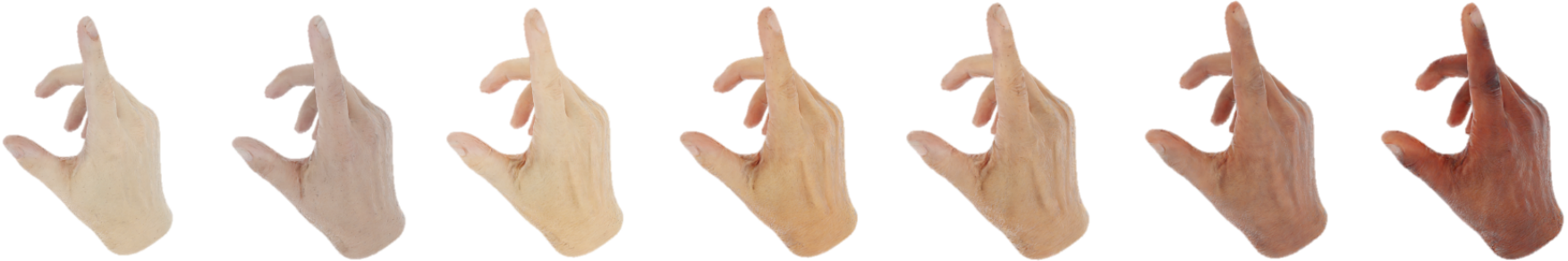}
  \caption{Same hand in different textures.}
  \label{fig:syn_tex}
\end{figure}

\noindent\textbf{Hand Texture.} We synthesize rich hand textures using the NIMBLE model; it is formulated by $\mathcal{A}(\alpha) = \bar{A} + \Phi\alpha$, where $\alpha$ is the appearance parameter, $\bar{A}$ is the average appearance and $\Phi$ are the model principal components. Existing synthetic hand datasets have unrealistic hand textures or limited texture diversity (\eg, RenderIH only offers 30 textures). NIMBLE hand textures are 
the linear interpolation of 38 photo-realistic hand texture assets from different ages, genders, and races to enhance texture diversity. Some examples are shown in Fig.~\ref{fig:syn_tex}.

\begin{figure}[tb]
  \centering
  \includegraphics[width=1\linewidth]{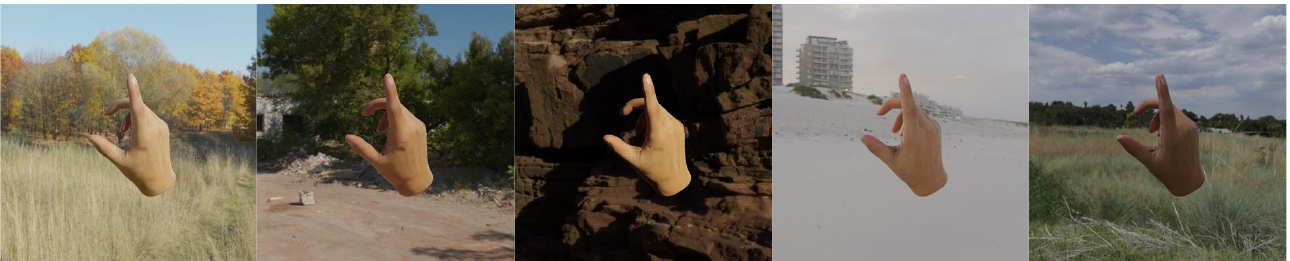}
  \caption{Same hand and hand texture in different backgrounds and lighting. The same hand looks different under different backgrounds and lighting.}
  \label{fig:syn_bg}
\end{figure}

\noindent\textbf{Background and Lighting.} We use 669 HDRI scenes from \cite{zaal_hdrihaven_2020} as the background; these images feature realistic indoor and outdoor scenes and have been used by \cite{black2023bedlam, li2023renderih, moon2023reinterhand, tse2023spectral} to provide highly photo-realistic rendered scenes. Some examples are shown in Fig.~\ref{fig:syn_bg}.

\noindent\textbf{Rendering.} 
We render with Blender Python based on the ray-tracing rendering engine Cycles \cite{blender_cycles_2021}.  
First, we import the skin mesh and apply texture to the mesh with diffuse, specular, and normal 
maps. The camera is set based on the target real datasets' camera intrinsics when analyzing. Backgrounds and textures are selected randomly.
This setup creates an automated rendering pipeline contained within Python with minimal effort.
It takes around 1 second to render an image with a single RTX A5000 GPU, with further parallelization speedups. 
More details can be found in Supplementary~Sec.~\tb{C}.

\begin{figure}[tb]
  \centering
  \includegraphics[width=1\linewidth]{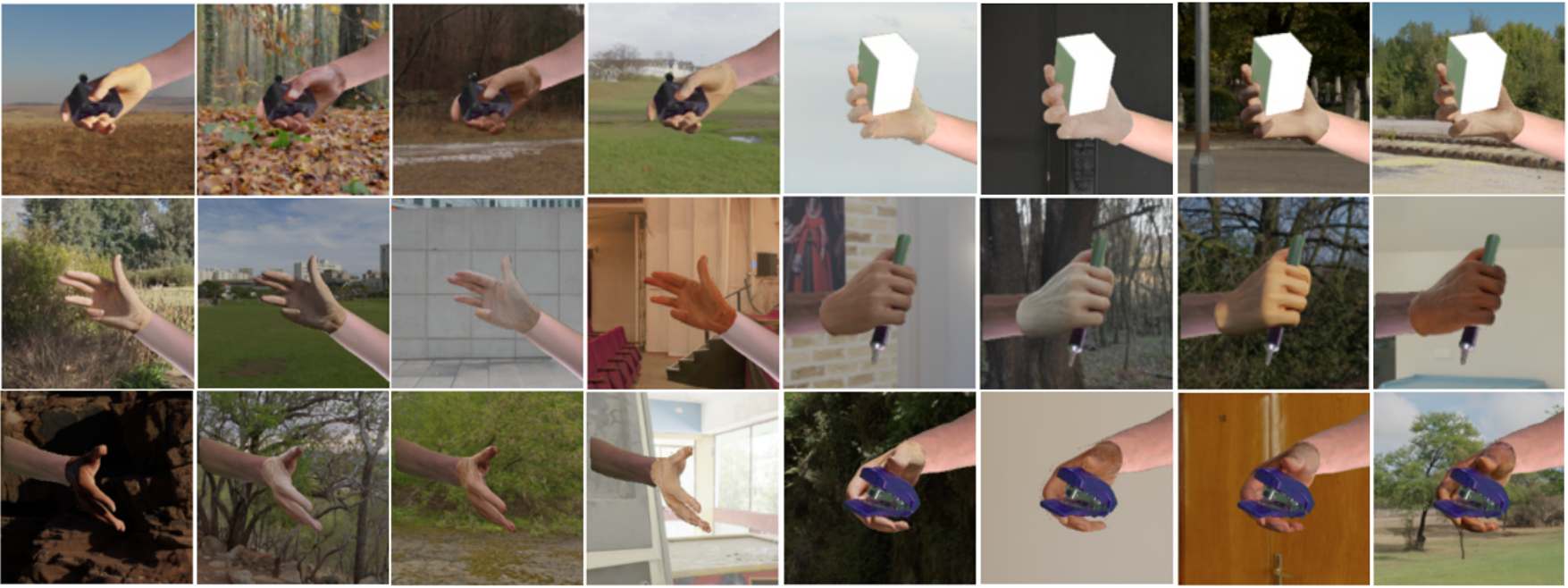}
  \caption{Synthetic hand images with realistic hand textures, backgrounds, lighting, arms, and interacting objects.}
  \label{fig:syn_dataset}
\end{figure}

\subsection{ Decomposition and Composition}\label{sec:hand_decompose}
 
Since our work focuses on analyzing the importance of different components, we segment the arm and object from real images and compose them into synthetic images, as shown in Fig.~\ref{fig:header}. The composition is more practical and serves as a better control case than rendering these components.

To segment the arm and object from real images,  
we use the pre-trained object detector Grounding DINO~\cite{liu2023grounding} to extract bounding boxes for these components. The bounding box crops are then fed into the segmentation network SAM~\cite{kirillov2023segment} to estimate the corresponding segmentation masks.
For each real image $I_{real}^i$, where $i \in [1, N]$, $N$ is the number of poses in the target real dataset, we extract the bounding boxes $bbox^i$ of the arm or interacting object with input text prompts $x$ by object detector $D(\cdot)$:  $bbox^i = D(x)$. Then the bounding boxes $bbox^i$ are passed to SAM $S(\cdot)$ to obtain the corresponding masks $M^i$: $M^i = S(bbox^i)$.

The final synthetic image $\tilde{I}_{syn}^j$ with arm and interacting object can be defined as follows:
\begin{align}
    \tilde{I}_{\text{syn}}^j &= (1 - M_{\text{obj}}^i - M_{\text{arm}}^i) \odot I_{\text{syn}}^j \nonumber \\
    &\quad + M_{\text{obj}}^i \odot I_{\text{real}}^i + M_{\text{arm}}^i \odot I_{\text{real}}^i,
\end{align}
where $I_{syn}^j$ denotes the rendered synthetic image, $M_{obj}^i$ and $M_{arm}^i$ denote the extracted object mask and arm mask, $\odot$ denotes element-wise multiplication and $j = k * N + i$, where $k \in \mathbb Z_{\ge 0}$. In this way, we get the final synthetic images. Examples are shown in Fig.~\ref{fig:syn_dataset}.

\subsection{Object Occlusion Levels and Priors}\label{sec:occ_prior}

\begin{figure}[tb]
  \centering
  \includegraphics[width=1\linewidth]{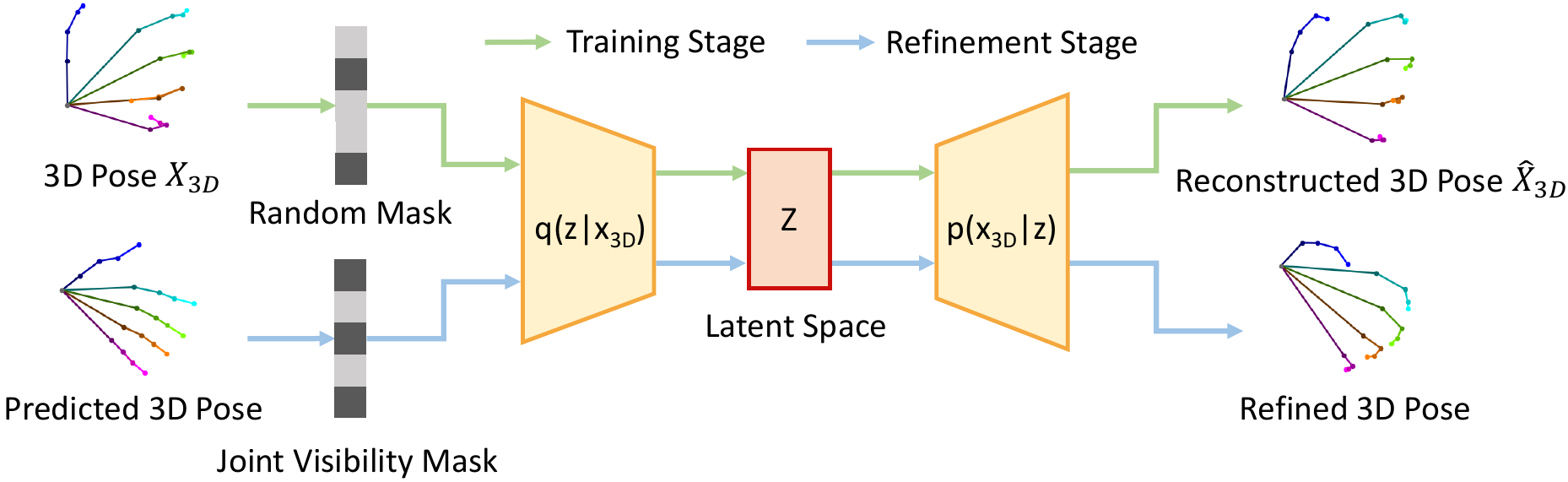}
  \caption{We build a VAE object occlusion prior and use this prior to refine the predicted hand poses.}
  \label{fig:occ_vae}
\end{figure}

To analyze the influence of object occlusion in synthetic-to-real adaptation, we divide the validation set into different levels of occlusion using~\cite{xu2023h2onet}, which provides finger-level occlusion labels (five fingers and palm) for each hand, ranging from 0 (least) to 6 (highest).
Moreover, to show the connection between occlusion levels and hand object priors, we employ a VAE prior to reconstruct occluded joints based on visible joints.
Similar to \cite{yang2019aligning, zuo2023reconstructing}, we use the VAE~\cite{kingma2013auto} to construct the object occlusion prior, which consists of an encoder $q(z|x_{3D})$ and a decoder $p(x_{3D}|z)$ (see Fig.~\ref{fig:occ_vae}). During the training stage, input 3D pose $x_{3D}$ is first randomly masked to increase reconstruction diversity. Then the encoder maps the masked 3D pose $x_{3D}$ to latent space $z$, and the decoder reconstructs a 3D pose $\hat{x}_{3D}$. A Kullback-Leibler divergence loss and a 3D joint loss are used to train the object occlusion prior:
\begin{equation}
    L_{VAE} = \lambda L_{KL} +  \| \hat{x}_{3D} - x_{3D} \|_2^2.
\end{equation}

During the refinement stage, we use the pre-trained VAE prior to refine the predicted joints. More details are provided in Secs.~\tb{D} and~\tb{E} of the Supplementary.

\section{Experiments}

\subsection{Datasets and Metrics}

We use FreiHAND and Dex-YCB as the target real datasets and synthesize their corresponding synthetic datasets, SynFrei and SynDex, for analysis.

\begin{itemize}

\item \textbf{FreiHAND}~\cite{zimmermann2019freihand} has 130k training and 4k testing images of a single hand. Each training sample features a real RGB image captured on a green screen background and three extra images with synthetic backgrounds.

\item \textbf{Dex-YCB}~\cite{chao2021dexycb} is a large-scale 3D hand-object dataset with over 582k images, featuring objects from the YCB dataset \cite{xiang2017posecnn}. We use the default ``S0'' train/test split and a reduced version of the training set, with 35k images for analysis.

\item \textbf{HO3D-v2}~\cite{hampali2020honnotate} features 3D hand-object interactions, with 
66k training and 11k testing images. The ten interacting objects used in the dataset are from the YCB dataset. We use HO3D-v2 to perform cross-dataset validation to evaluate the model's performance on samples with unseen interacting objects.

\item \textbf{MOW}~\cite{cao2021reconstructing} is a 3D dataset of humans manipulating objects in-the-wild, including 512 annotated samples. It is used as the test set to evaluate the generalization ability.

\end{itemize}

\noindent\textbf{Evaluation Metrics.} We evaluate hand pose and shape accuracy using standard metrics PA-MPJPE and PA-MPVPE in cm, which compare the mean per joint position error and mean per vertex position error of prediction and ground truth after Procrustes Alignment.

\subsection{Baselines and Training Details}
We reproduce five 3D hand pose estimation methods: S$^2$HAND \cite{chen2021model}, CMR \cite{bib:CMR}, METRO \cite{lin2021end-to-end}, MeshGraphormer \cite{lin2021mesh} and simpleHand \cite{zhou2024simple} on the real target datasets as baselines using their official training code. Additionally, we reproduce these methods on our synthesized datasets with amplitude spectrum augmentation~\cite{chattopadhyay2023pasta} (See Supplementary~Sec.~\tb{A}) for comparisons to demonstrate the effectiveness of our data synthesis pipeline.
All baselines are trained with one NVIDIA RTX A5000 GPU. We mainly use S$^2$HAND to analyze the synthetic-to-real gap.

\subsection{Comparisons with Baselines}
We train five methods on the real dataset FreiHAND as baselines and on our synthesized SynFrei to show the effectiveness of our synthetic data synthesis pipeline. As shown in Tab.~\ref{tab:app_baselines}, models trained with only our synthetic dataset can achieve state-of-the-art results, which achieve 97\%, 91\%, 88\%, 91\% and 84\% of the performance of models trained with the real dataset for S$^2$HAND, CMR, METRO, MeshGraphormer, and simpleHand respectively.
These results indicate a synthetic-to-real gap of around 0.1 cm. We notice decreased performance when training with only synthetic data, especially with stronger model backbones. We speculate that the stronger models over-fit more easily to the synthetic data, affecting the synthetic-to-real generalization performance.

We also train S$^2$HAND on the reduced version of Dex-YCB and our synthesized SynDex, achieving comparable performance on the Dex-YCB test set of 0.86 cm vs. 0.87 cm in PA-MPJPE.

\begin{table}[tb]
  \caption{Comparisons of methods trained on FreiHAND and only our SynFrei on the FreiHAND test set 
  (PA-MPJPE$\downarrow$ / PA-MPVPE$\downarrow$).}
  \label{tab:app_baselines}
  \centering
  \resizebox{1\linewidth}{!}{
    \begin{tabular}{l|ccc}
    \toprule
    Method & Real FreiHAND & Our SynFrei & Syn$\rightarrow$Real \\
    \midrule
     S$^2$HAND \cite{chen2021model} &  0.99 / 1.02 & 1.02 / 1.05 & 97\% \\
     CMR \cite{bib:CMR} & 0.77 / 0.78 & 0.85 / 0.88 & 91\% \\
     METRO \cite{lin2021end-to-end} & 0.69 / 0.71 & 0.78 / 0.79 & 88\% \\
     MeshGraphormer \cite{lin2021mesh}  & 0.69 / 0.70 & 0.76 / 0.78 & 91\% \\
     simpleHand \cite{zhou2024simple} & 0.65 / 0.66 & 0.77 / 0.79 & 84\% \\
  \bottomrule
  \end{tabular}
  }
\end{table}

\begin{figure*}[h]
  \centering
  \includegraphics[width=1\linewidth]{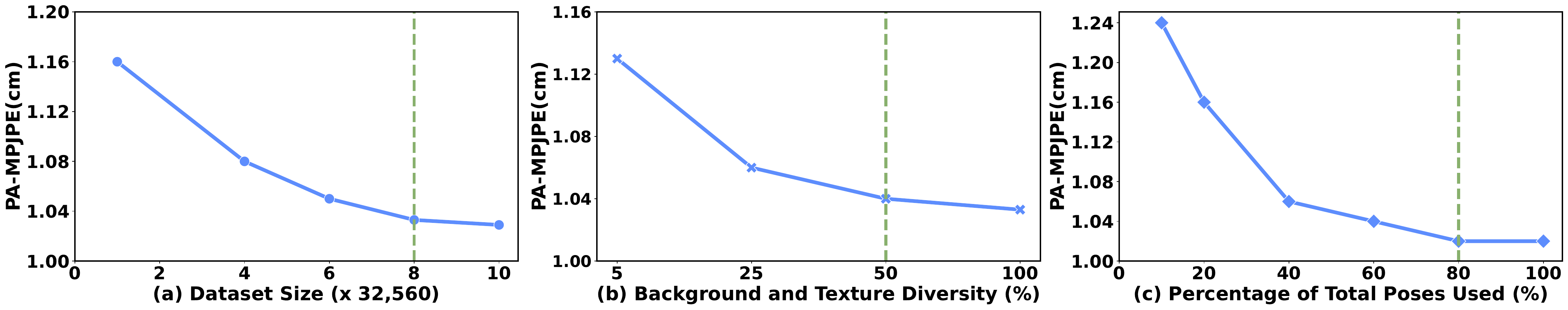}
  \caption{(a) Synthetic dataset size influence. (b) Background and hand texture diversity influence. (c) The percentage of real hand poses used for synthesizing data. The green lines indicate the saturation points.}
  \label{fig:data_size_bg_div}
\end{figure*}

\subsection{Comparison with Other Synthetic Datasets}
We compare with DARTset~\cite{gao2022dart}, which has a more diverse pose distribution, hand wrists, and accessories but lacks dynamic backgrounds, interacting objects, and amplitude spectrum augmentation. DARTset samples their hand poses from FreiHAND and augments them by linear interpolating the synthesized pose from A-MANO space and FreiHAND sample pose. As shown in Tab.~\ref{tab:app_dart}, CMR and METRO trained with our SynFrei outperform DARTset despite our dataset size being half DARTset (325,600 vs. 758,378). Moreover, CMR trained with our SynDex outperforms DARTset on the FreiHand test set (1.92 / 1.98), even with different pose distributions.

Comparisons between single- and two-hand datasets are challenging due to the domain gap. We still evaluate the cross-dataset generalization on the HO3D-v2 test set for models trained with SynFrei and RenderIH~\cite{li2023renderih}. We only take the right-hand results of the RenderIH model. Models trained with SynFrei achieve better cross-dataset evaluation than RenderIH of 1.41 cm vs. 1.76 cm in PA-MPJPE. This result can be attributed to the limited expressiveness of interacting hands, which fails to let the model learn hand-object interactions, and the blurry background effects in RenderIH images due to the low background resolution.

\begin{table}[tb]
  \caption{Comparisons of methods trained on DARTset and our SynFrei on the FreiHAND test set (PA-MPJPE$\downarrow$ / PA-MPVPE$\downarrow$). The results of DARTset are from its paper.}
  \label{tab:app_dart}
  \centering
  \resizebox{0.6\linewidth}{!}{
  \begin{tabular}{l|cc}
    \toprule
    Method & DARTset & SynFrei \\
    \midrule
     CMR &  2.56 / 2.57 & \textbf{0.85} / \textbf{0.88} \\  
    METRO &  1.77 / 1.75 & \textbf{0.78} / \textbf{0.79} \\ 
    \bottomrule
  \end{tabular}
  }
\end{table}

\begin{table}[t]
  \caption{Arm, amplitude spectrum augmentation, and interacting object influence on FreiHAND and Dex-YCB test sets (PA-MPJPE$\downarrow$). ``Amp Aug" is short for amplitude spectrum augmentation. \halfcheckmark refers to assigning RGB values to the arm and object mask positions.}
  \label{tab:app_arm_aug}
  \centering
  \resizebox{1\linewidth}{!}{
  \begin{tabular}{l|ccc|cc}
    \toprule
    & \multicolumn{3}{c|}{\textbf{Components}} & \multicolumn{2}{c}{\textbf{Train set}} \\
    & Arm & Amp Aug & Object & SynFrei & SynDex \\
    \midrule
   (i) & & \newcheckmark & \newcheckmark  & 1.07 & 0.90 \\
   (ii)  & \newcheckmark &  & \newcheckmark & 1.11 & 0.89 \\
   (iii) & \newcheckmark & \newcheckmark &  & 1.07 & 0.95 \\
   (iv)  & \halfcheckmark & \newcheckmark & \halfcheckmark & 1.04 & 0.92 \\
   (v)  & \newcheckmark & \newcheckmark & \newcheckmark  & \textbf{1.02} & \textbf{0.87} \\
   \bottomrule
   \end{tabular}
   }
\end{table}

\begin{figure*}[t]
  \centering
  \includegraphics[width=1\linewidth]{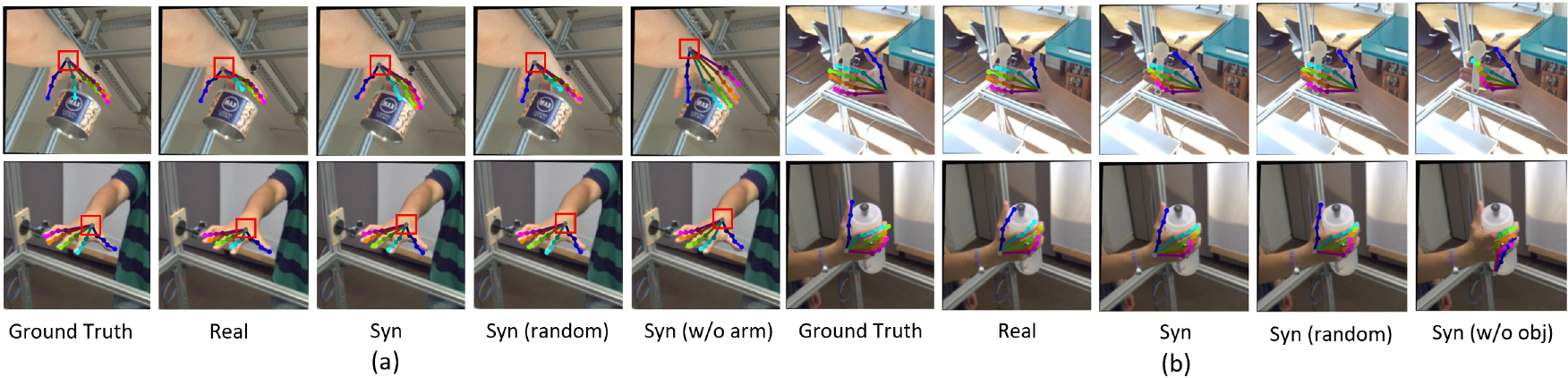}
  \caption{(a) Comparisons of model trained with FreiHAND (Real), our SynFrei (Syn), our SynFrei with random arm (random), and our SynFrei (w/o arm) on FreiHAND test set. (b) Comparisons of model trained with FreiHAND, SynFrei, SynFrei with random interacting object and SynFrei (w/o object) on the FreiHAND test set.}
  \label{fig:arm_obj_abla}
\end{figure*}

\subsection{Appearance Analysis}

\noindent\textbf{Dataset Size and Diversity Influence.}\label{sec:app_size}
We analyze the influence of the synthetic dataset size on synthetic-to-real performance. As illustrated in Fig.~\ref{fig:data_size_bg_div}(a) and, as expected, there are diminishing returns with adding more data.  
When SynFrei's synthetic dataset size reaches 8$\times$32,560, performance gains plateau and the trend of error decrease becomes flat.

We further render synthetic datasets at this full size of 8$\times$32,560 samples, but with varying levels of diversity: 50\%, 25\%, and 5\%, where each value indicates the percentage of the full background and hand texture assets. As shown in Fig.~\ref{fig:data_size_bg_div}(b), more diverse backgrounds and hand textures can reduce the synthetic-to-real gap. This result suggests that enriching the background and texture assets is useful when synthesizing data. However, the performance plateaus at 50\%, indicating that around 300 indoor and outdoor HDRI scenes are sufficient for synthesizing hand data. Several techniques can be implemented in our synthesis pipeline to further improve performance, such as changing camera viewpoints, flipping or rotating backgrounds, and adjusting skin tone contrast, brightness, or subsurface color.

\noindent\textbf{Arm Influence.}\label{sec:app_arm_aug}
We analyze the importance of the arm by removing it from synthetic images. Tab.~\ref{tab:app_arm_aug} shows increases in PA-MPJPE for models trained with SynFrei and SynDex without the arm component comparing Rows (i) and (v). Fig.~\ref{fig:arm_obj_abla}(a) shows that the model trained on SynFrei without the arm tends to misinterpret parts of the arm as the wrist, especially if the arm occupies a large portion of the image. Such errors do not occur for SynFrei with arms. Since the arm and hand share similar colors, models can easily be confused and incorrectly locate the root joint on the arm rather than the wrist if synthetic images lack arm information. This result suggests that the arm is an important cue for locating the wrist, thus affecting the overall hand pose.

\noindent\textbf{Amplitude Spectrum Augmentation Influence.}\label{sec:app_amp_aug}
We find that amplitude spectrum augmentation is a crucial component. Removing it results in increases in PA-MPJPE for both SynFrei (0.09 cm) and SynDex (0.02 cm) (see Tab. \ref{tab:app_arm_aug} Rows (ii) and (v)). This suggests that the ability of amplitude spectrum augmentation to distort amplitude information while preserving phase information (\eg, hand structure) can enhance model robustness when the domain shifts from synthetic to real.

\noindent\textbf{Random Arm and Object Occlusion Influence.}
We further randomize the arm and object incorporated into the hand image by assigning RGB values to the corresponding arm and object mask positions described in Sec.~\ref{sec:hand_decompose}. For the arm, we assign it to the mean RGB value of the hand to better imitate hand-arm compatibility. For the object occlusion, we assign a random RGB value to the object position. Tab.~\ref{tab:app_arm_aug} Rows (iv) and (v) show that with random arm and object occlusion, although the synthetic-to-real gaps slightly increase compared with segmenting from real images, the synthetic-to-real performance can still achieve 95\% and 93\% on FreiHAND and Dex-YCB test sets. This result reveals that the model can learn how to locate the wrist position and interacting objects accurately without highly realistic arm or object occlusion information (see Fig.~\ref{fig:arm_obj_abla}).

\subsection{Pose Analysis}

We analyze the influence of hand pose on the synthetic-to-real performance by controlling the number of real hand poses used for data synthesis. Note that we still diversify each hand with different backgrounds and textures for rendering. 
As shown in Fig.~\ref{fig:data_size_bg_div}(c), the synthetic-to-real performance improves when synthetic hand poses approach the real ones. This result suggests that the pose difference between the synthetic and real datasets introduces a synthetic-to-real gap. We also observe that the model trained with synthetic data can achieve around 80\% of the final performance using only 5\% of total real poses. With 20\% of real poses, the performance increases to around 90\%, and with 40\% of real poses, it reaches 97\%. In other words, synthetic hand images effectively enable the model to learn visual representations, while core hand poses are sufficient to achieve good generalization performance.

\subsection{Object Occlusion Analysis}

Tab.~\ref{tab:app_arm_aug} Rows (iii) and (v) show increases in PA-MPJPE after removing the interacting objects from SynFrei (0.05 cm) and SynDex (0.08 cm), especially for the Dex-YCB test set, which features more hand-object interactions. When the object largely occludes the hand, the predicted hand pose does not interact with the object right after removing interacting objects from synthetic images (see Fig.~\ref{fig:arm_obj_abla}(b)).

\noindent\textbf{Occlusion Levels.}\label{sec:occ_cross}
We use the HO3D-v2 dataset for cross-dataset validation, which features hand-object interactions. Since HO3D-v2 test set labels are not publicly available, we separate the original train set into train and validation sets.

\begin{figure}[tb]
  \centering
    \includegraphics[width=1\linewidth]{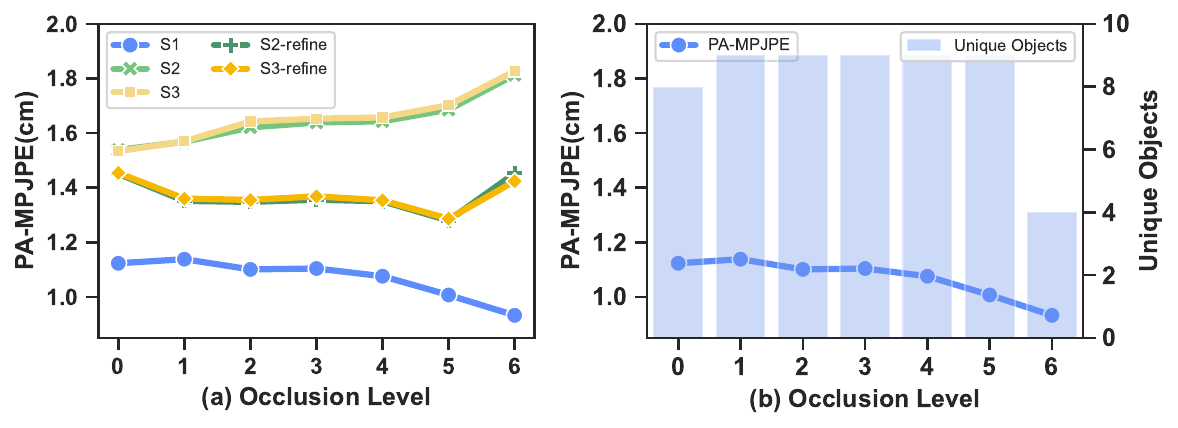}
  \caption{(a) Comparisons of $S_1$, $S_2$ and $S_3$ on validation sets with different occlusion levels. $S_2$ and $S_3$ after object occlusion prior refinement are denoted as $S_2$-refine and $S_3$-refine.
  (b) Relationship between the performance of $S_1$ and the number of unique objects at each occlusion level.
  }
  \label{fig:occ_vae_refine}
\end{figure}

We compare models' performance on validation sets under three training settings: S$^2$HAND trained with HO3D-v2 train set ($S_1$), FreiHAND ($S_2$), and SynFrei ($S_3$). As shown in Fig.~\ref{fig:occ_vae_refine}(a), a performance gap exists between $S_1$ and $S_2$, $S_3$. As the occlusion level rises, $S_2$ and $S_3$ degrade due to their lack of object occlusion information in the validation sets, while $S_1$ improves, which incorporates this prior.

We further investigate the distribution of unique object amount across each occlusion level and discover a relationship between the performance of $S_1$ and the unique object amount. As shown in Fig.~\ref{fig:occ_vae_refine}(b), $S_1$ performs the best on occlusion level 6, which has the least amount of unique objects. With object occlusion prior information in validation sets, $S_1$ can associate hand poses with specific objects. Also, higher occlusion levels are often associated with larger objects, which impose greater constraints on hand poses and result in smaller variations in poses. Therefore, $S_1$ demonstrates improved performance on higher occlusion levels. However, without object occlusion information, the performances of $S_2$ and $S_3$ decline as the occlusion levels increase. These findings show that object occlusion prior contributes to the synthetic-to-real gap.
 
\noindent\textbf{Object Occlusion Prior.}\label{sec:occ_vae}
We use Dex-YCB to train a VAE object occlusion prior (See Sec.~\ref{sec:occ_prior}), which includes the objects in HO3D-v2. As shown in Fig.~\ref{fig:occ_vae_refine}(a), the synthetic-to-real performance of $S_3$ improves after applying the VAE object occlusion prior to refine the predicted joints ($S_3$-refine). 
Incorporating the object occlusion prior bridges the gap between synthetic and real hand datasets with high levels of occlusion. While previous works use prior information to refine poses in real-to-real scenarios~\cite{nam2023cyclic, zuo2023reconstructing, rempe2021humor}, our findings also show its utility in synthetic-to-real adaptation.

Through our object occlusion analysis, we provide a potential training scheme to better adapt synthetic data to real applications. Models can initially be trained on large-scale synthetic data to learn hand representations. Existing hand-object interaction datasets can then be used to provide useful prior information to refine the predictions and further reduce the synthetic-to-real gap.

\subsection{Hand Skeleton Topology Analysis}
Different hand skeleton topologies can have different joint definitions \cite{9262071}. This can introduce a synthetic-to-real gap since the synthetic data's skeleton topology may differ from real datasets.
To analyze the influence of hand skeleton topology difference, we train S$^2$HAND without label adaption and obtain NIMBLE joints as the ground-truth joints by fitting NIMBLE mesh to MANO mesh for training. Note that the NIMBLE skeleton differs from the MANO skeleton (see Supp.~\tb{F}), and we remove the extra anatomical joints of NIMBLE when evaluating. As shown in Tab.~\ref{tab:tem}, training with NIMBLE joints and without label adaption, the synthetic-to-real error increases from 1.02 cm to 1.28 cm. This result demonstrates that differences in hand skeleton topology between synthetic and real datasets can introduce bias to the prediction.

\begin{table}[t]
  \caption{Hand skeleton topology differences.}
  \label{tab:tem}
  \centering
  \scalebox{0.9}{
  \begin{tabular}{c|c}
    \toprule
    Label Adaptation & PA-MPJPE$\downarrow$ \\
    \midrule
     - & 1.28\\
     \newcheckmark & \textbf{1.02} \\
    \bottomrule
  \end{tabular}
  }
\end{table}

\subsection{Cross-Dataset Evaluation}
We demonstrate the ability of our synthetic data to bridge the gap across different hand datasets through cross-dataset generalization using S$^2$HAND. As shown in Tab.~\ref{tab:cross}, the model trained with a mixture of real and synthetic data achieves better results than only real data for in- and cross-domain generalization. For example, the model trained with Dex-YCB and SynDex performs better than with only Dex-YCB on the FreiHAND test set. The difference may be because Dex-YCB mostly consists of indoor images, while the FreiHAND test set contains many outdoor images. Incorporating synthetic images with both indoor and outdoor environments bridges this gap. These results prove our synthetic data is a good complement to real data, both in- and cross-domain. More results can be found in Supplementary~Sec.~\tb{H}.

\begin{table}[t]
  \caption{Cross-dataset generalization results in PA-MPJPE $\downarrow$.
  $^\dag$: results come from the official evaluation server.}
  \label{tab:cross}
  \centering
  \resizebox{1\linewidth}{!}{
  \begin{tabular}{l|ccccc}
    \toprule
    \diagbox[height=2.5em,width=8em]{Train Set}{Test Set} & FreiHAND & Dex-YCB  & HO3D-v2$^\dag$ & RHD & MOW \\
    \midrule
     FreiHAND & 0.99 & 1.14 & 1.39 & 1.40 & 1.47 \\
     ~ + SynFrei & \textbf{0.95} & \textbf{1.11} & \textbf{1.28} & \textbf{1.32} & \textbf{1.38} \\
     \midrule
     Dex-YCB & 1.63 & 0.86 & 1.25 & 1.72 & 1.24 \\
     ~ + SynDex & \textbf{1.49} & \textbf{0.82} & \textbf{1.15} & \textbf{1.65} & \textbf{1.18} \\
    \bottomrule
  \end{tabular}
  }
\end{table}

\section{Conclusion}

This paper proposes the first systematic study of the synthetic-to-real gap of 3D hand pose estimation. We analyze the synthetic-to-real gap with respect to appearance, hand pose, object occlusion, and hand skeleton topology. We impressively demonstrate that synthetic hand data can achieve the same level of accuracy as real hand data while paving the path to use synthetic data alone. Models trained on a mixture of real and our synthetic data outperform those trained only on real data for in-domain and cross-domain generalization. Our proposed data synthesis pipeline can create high-quality synthetic hand images to alleviate our reliance on real data and benefit future research and applications.

\section*{Acknowledgements}
This research / project is supported by the Ministry of Education, Singapore, under the Academic Research Fund Tier 1 (FY2022). Additionally, this work is supported by the National Natural Science Foundation of China (No. 62406298), the Guangzhou Municipal Nansha District Science and Technology Bureau (No.2022ZD012),  the HSBC project L0562, and the Fundamental Research Funds for the Central Universities, Communication University of China (CUC24XT07).

{
    \small
    \bibliographystyle{ieeenat_fullname}
    \bibliography{main}
}

\setcounter{figure}{0}
\renewcommand\thetable{\Alph{table}} 
\setcounter{table}{0}
\renewcommand{\thesection}{\Alph{section}}
\setcounter{section}{0}
\setcounter{page}{1}
\maketitlesupplementary

In the following supplementary material, we present more details about our work. Sec.~\textcolor{cvprblue}{A} provides the amplitude spectrum augmentation algorithm used during the model training stage to augment synthetic images. Sec.~\textcolor{cvprblue}{B} provides the fitting algorithm for obtaining NIMBLE hand meshes in different poses. Sec.~\textcolor{cvprblue}{C} provides the rendering details of the hand image synthesis pipeline. Sec.~\textcolor{cvprblue}{D} illustrates how we obtain the finger-level occlusion annotations for object occlusion analysis. Sec.~\textcolor{cvprblue}{E} provides the implementation details of the VAE object occlusion prior. Sec.~\textcolor{cvprblue}{F} shows the difference between NIMBLE and MANO hand templates. Sec.~\textcolor{cvprblue}{G} compares the influence of hand model templates used for data synthesis. Sec.~\textcolor{cvprblue}{H} presents the results of training the model with a combination of synthetic and real data. Sec.~\textcolor{cvprblue}{I} shows the component influence before Procrustes Alignment. Sec.~\textcolor{cvprblue}{J} shows the comparison of synthetic data generated by generative models. Sec.~\textcolor{cvprblue}{K} shows more visualization of our synthetic dataset.

\section{Amplitude Spectrum Augmentation}\label{supp:A}
We apply amplitude spectrum augmentation~\cite{chattopadhyay2023pasta} during the model training stage to address the frequency domain gap. 
The amplitude spectrum augmentation algorithm is shown in Alg.~\ref{alg:amp}. We set the hyper-parameters $\alpha$ to 3, $\beta$ to 0.25, and $k$ to 2 during the training.

\begin{algorithm}[H]
    \caption{Amplitude spectrum augmentation}
    \label{alg:amp}
    \begin{algorithmic}[1]
        \STATE Input: $x$ $\in$ $\mathbb{R}^{H \times W}$  \Comment*[r]{\textnormal{Input synthetic image}}
        \STATE Augmentation parameters: $\alpha, k, \beta$
        \STATE $\mathcal{F}(x) \leftarrow \textnormal{FFT($x$)}$ \Comment*[r]{\textnormal{Fast Fourier Transform}}
        \STATE $\mathcal{A}(x), \mathcal{P}(x) \leftarrow \textnormal{Abs}(\mathcal{F}(x)), \textnormal{Ang}(\mathcal{F}(x))$ %
        \STATE $\mathcal{A}(x) \leftarrow \textnormal{FFTShift}(\mathcal{A}(x))$ \Comment*[r]{\textnormal{Zero-center amplitude}}
        \FOR{$p$ $\in$ [$-H / 2$, $H / 2$]}
            \FOR{$q$ $\in$ [$-W / 2$, $W / 2$]}
            \STATE $\sigma [p, q] \leftarrow \left(2\alpha\sqrt{\frac{p^2 + q^2}{H^2 + W^2}}\right)^k + \beta$
            \ENDFOR
        \ENDFOR
        \STATE $\lambda \sim \mathcal{N}(1, \sigma^2)$ \Comment*[r]{\textnormal{Calculate perturbations}}
        \STATE $\tilde{\mathcal{A}}(x) \leftarrow \lambda \odot \mathcal{A}(x)$  \Comment*[r]{\textnormal{Perturb amplitude spectrum}}
        \STATE $\tilde{\mathcal{A}}(x) \leftarrow \textnormal{FFTShift}(\tilde{\mathcal{A}}(x))$
        \STATE $\tilde{x} \leftarrow \textnormal{Inverse-FFT}(\tilde{\mathcal{A}}(x), \mathcal{P}(x))$  \Comment*[r]{\textnormal{Augmented image}}
    \end{algorithmic}
\end{algorithm}

\section{Fitting Algorithm}\label{B}
\label{sec:B}
To prepare hand meshes for rendering, we develop algorithms to fit NIMBLE~\cite{li2022nimble} meshes to MANO~\cite{romero2022embodied} meshes. Since we find that optimizing all parameters directly leads to poor fitting results, we divide the fitting process into two stages. In the coarse stage, we optimize the rotation parameter $r \in \mathbb{R}^3$ and translation parameter $t \in \mathbb{R}^3$. The training loss consists of a 3D joint loss and a vertex loss:
\[ L = L_{joint} + \lambda_{vert} L_{vert},\]
where $\lambda_{vert}$ is set to 0.1. We use a learning rate of 1e0 for both $r$ and $t$. The epoch for the coarse stage is 2, with 3,000 iterations per epoch.

In the fine stage, we optimize pose $\theta \in \mathbb{R}^{30}$, shape $\beta \in \mathbb{R}^{20}$, rotation $r$ and translation $t$ parameters together. The training loss consists of a vertex loss and regularization losses for pose and shape parameters:
\[L = L_{vert} + \lambda_{pose} L_{pose} + \lambda_{shape} L_{shape},\]
where $\lambda_{pose}$ and $\lambda_{shape}$ are set to 50. We use a learning rate of 1e-3 for both $\theta$ and $\beta$, and a learning rate of 1e-2 for both $r$ and $t$. The epoch for the fine stage is 4, with 3,000 iterations per epoch.
All learning rates are divided by 10 every 1,000 iterations. Additionally, learning rates are reset to their initial value at the beginning of each epoch to prevent convergence to local minima.

\begin{figure}[tb]
  \centering
  \includegraphics[width=1\linewidth]{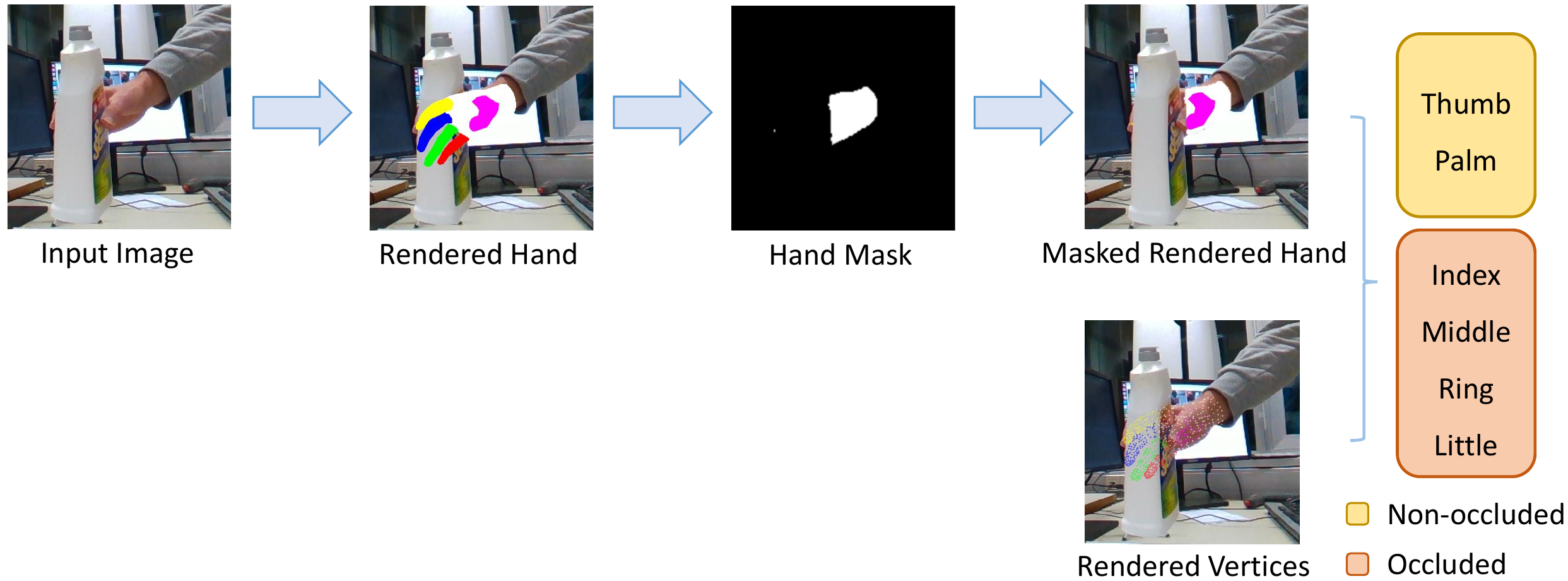}
  \caption{Finger-level occlusion annotation preparation.}
  \label{fig:occ_pre}
\end{figure}

\section{Rendering Details}\label{C}
In this section, we detail our camera settings for rendering. To ensure our analysis is not affected by other influencing factors, our camera viewpoints are consistent with the target real dataset we compared. We calculate the camera focal length $f^i$ for rendering each synthetic image $i$ by:

\[ f^i = f_x^i * \frac{W_{sensor}}{W_{image}},\]
where $f_x^i$ is derived from the provided camera intrinsic matrix $K^i$, $W_{sensor}$ represents the sensor width, which is 36 mm, and $W_{image}$ represents the width of the rendered image in pixel, which can be 224 or 256.
Multi-view settings can also be implemented using our rendering code.

\section{Occlusion Level Division}\label{D}
We introduce how we collect the finger-level occlusion annotations using \cite{xu2023h2onet}. As shown in Fig.~\ref{fig:occ_pre}, we first allocate faces and vertices of five fingers and the palm with different colors and then render the hand and vertices into the input image. Since HO3D-v2~\cite{hampali2020honnotate} provides the hand mask for each image, we apply this mask to remove the rendered hand parts occluded by the object. With the masked rendered hand and rendered vertices, we can compare each vertex color against the corresponding finger color to calculate the number of occluded vertices. Here, we set the threshold of occluded vertices to 40 to determine whether the finger can be considered occluded. With these occlusion annotations, we can divide the validation set into different occlusion levels and perform cross-dataset validation.

\begin{table}[tb]
\centering
\begin{minipage}{.45\linewidth}
  \centering
    \centering
    \begin{tabular}{|c|c|}
    \hline
    \multicolumn{2}{|c|}{Encoder}  \\
    \hline
    Linear(512) & ReLU  \\
    \hline
    Linear(512) & ReLU  \\
    \hline
    Linear(512) & ReLU \\
    \hline
    \multicolumn{2}{|c|}{Linear(512)}  \\   
    \hline
    \end{tabular}
    \caption{VAE encoder.}
    \label{tab:encoder}
\end{minipage}%
\begin{minipage}{.45\linewidth}
  \centering
  \begin{tabular}{|c|c|}
    \hline
    \multicolumn{2}{|c|}{Decoder}  \\
    \hline
    Linear(512) & ReLU  \\
    \hline
    Linear(512) & ReLU  \\
    \hline
    Linear(512) & ReLU \\
    \hline
    Linear(512) & ReLU \\
    \hline
    Linear(512) & ReLU \\
    \hline
    \multicolumn{2}{|c|}{Linear(512)}  \\   
    \hline
    \end{tabular}
  \caption{VAE decoder.}
  \label{tab:decoder}
\end{minipage}
\end{table}

\section{VAE Architecture}\label{E}
The architecture of VAE object occlusion prior with an encoder and a decoder consists of a series of (Linear, ReLU) layers (see Tab.~\ref{tab:encoder} and Tab.~\ref{tab:decoder}). The dimension of latent variable $z$ is set to 64. The batch size is 128 and the learning rate is 1e-4. The total epoch for training the VAE object occlusion prior is 160. $\lambda$ is set to 0.01. 
In the training stage, a Kullback-Leibler divergence loss and a 3D joint loss are used to train the object occlusion prior and the random mask is applied to the input 3D pose. In the refinement stage, we apply the visibility mask to the predicted 3D pose and use the object occlusion prior to reconstruct those occluded joints. The visibility masks are obtained through a render-and-compare method similar to the finger-level occlusion annotations~\cite{xu2023h2onet}.

\section{Hand Skeleton Topology Difference}\label{F}
\begin{figure}[h]
  \centering
  \includegraphics[width=0.7\linewidth]{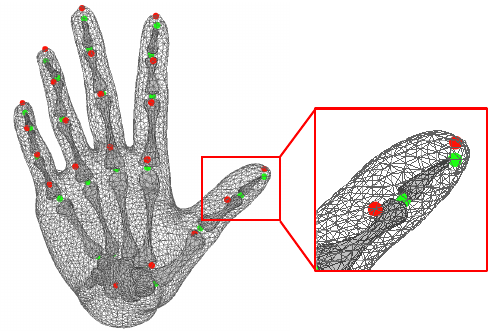}
  \caption{Hand skeleton topology difference between NIMBLE template and MANO template. The green dot denotes the NIMBLE joints and the red dot denotes the MANO joints.}
  \label{fig:tem}
\end{figure}

The hand skeleton topology difference between NIMBLE and MANO is shown in Fig.~\ref{fig:tem}. Different hand skeleton topologies can have different joint rig definitions and this difference can introduce bias to synthetic-to-real adaptation.

\section{NIMBLE vs. MANO}\label{G}

We render 10k synthetic images in NIMBLE and MANO models respectively. We keep other influencing factors consistent, such as poses and background diversity. The MANO texture maps are collected from RenderIH~\cite{li2023renderih}. As shown in Tab.~\ref{tab:template}, using NIMBLE achieves better performance than MANO across all metrics. This suggests that leveraging a finer hand mesh template for image synthesis is beneficial for bridging the synthetic-to-real gap.

\begin{table}[h]
  \caption{Comparison of S$^2$HAND trained with 10k synthetic images in MANO and NIMBLE template and tested on FreiHAND}.
  \label{tab:template}
  \centering
  \resizebox{1\linewidth}{!}{
  \begin{tabular}{l|cccc}
      \toprule
       Template & PA-MPJPE $\downarrow$ & PA-MPVPE $\downarrow$ & MPJPE $\downarrow$ & MPVPE $\downarrow$ \\
      \midrule
       MANO & 1.29 & 1.31 & 3.79 & 3.95 \\
       NIMBLE & \textbf{1.24} & \textbf{1.26} & \textbf{2.95} & \textbf{3.06} \\
      \bottomrule
  \end{tabular}
  }
\end{table}

\section{Synthetic-to-Real Adaptation}\label{H}

\begin{figure}[h]
    \centering
    \begin{subfigure}[b]{0.7\linewidth}{
        \centering
        \includegraphics[width=1\linewidth]{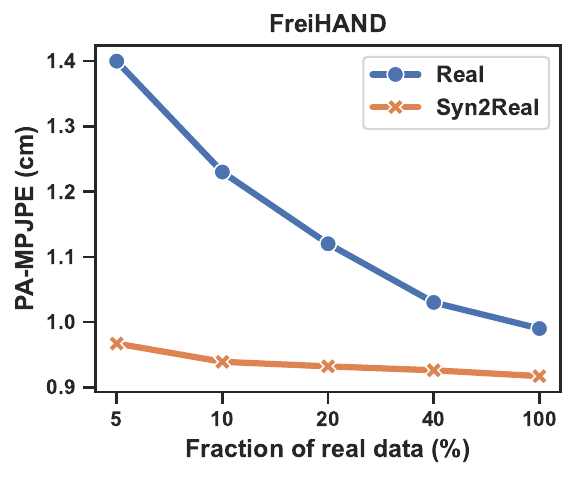}
    }
    \end{subfigure}
    \begin{subfigure}[b]{0.7\linewidth}{
        \centering
        \includegraphics[width=1\linewidth]{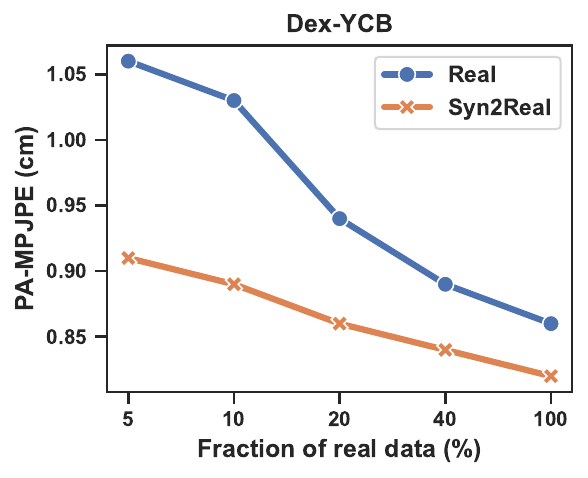}
    }
    \end{subfigure}
    \caption{Comparison of training with real data only and pre-training with total synthetic data followed by fine-tuning on real data (Syn2Real).}
    \label{fig:combined}
\end{figure}

We explore the advantages of pre-training on our total synthetic data when targeting FreiHAND and Dex-YCB. As shown in Fig.~\ref{fig:combined}, by leveraging our synthetic data for pre-training, the model achieves good performance even when fine-tuned with only a small fraction of real data.

\begin{figure}[h]
  \centering
  \includegraphics[width=1\linewidth]{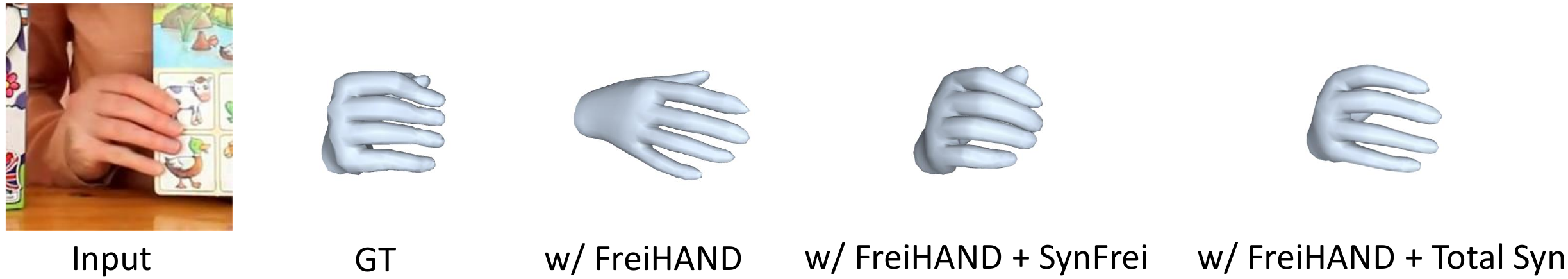}
  \caption{Qualitative comparison on MOW.}
  \label{fig:mow}
\end{figure}

Furthermore, in the in-the-wild scenario, synthetic data is useful as well. As shown in Tab.~\ref{tab:mow}, complementing real datasets with our total synthetic data achieves better performance on MOW. A qualitative comparison is presented in Fig.~\ref{fig:mow}.

\begin{table}[h]
\vspace{-1mm}
  \caption{Cross-dataset evaluation of S$^2$HAND on in-the-wild dataset MOW.}
  \label{tab:mow}
  \centering
\vspace{-3mm}
  \resizebox{1\linewidth}{!}{
  \begin{tabular}{lc|lc}
      \toprule
      Train sets & PA-MPJPE $\downarrow$ & Train sets & PA-MPJPE $\downarrow$  \\
      \midrule
       FreiHAND  & 1.47 & Dex-YCB & 1.24 \\
       + SynFrei & 1.38 & + SynDex & 1.18  \\
       + Total Syn Data & \textbf{1.19} & + Total Syn Data & \textbf{1.15} \\
      \bottomrule
  \end{tabular}
  }
\end{table}

\section{Component Influence before Procrustes Alignment}\label{I}
In this section, we show how each component influences the final predictions in metrics before Procrustes Alignment (MPJPE). As shown in Tab.~\ref{tab:metric}, adding different components has greater impacts on MPJPE than PA-MPJPE. For example, the improvement in MPJPE is greater than PA-MPJPE (12\%~vs.~5\%) for the arm. This confirms arm is important to locate the wrist and global rotation. Moreover, amplitude spectrum augmentation is crucial to reduce the synthetic-to-real gap. Removing it during training leads to a 22\% performance decline in MPJPE.

\begin{table}[t]
\vspace{-1mm}
\caption{Before vs. after Procrustes Alignment on FreiHAND test set across different factors. \halfcheckmark refers to assigning RGB values to the arm and object mask positions.
}
\vspace{-3mm}
  \label{tab:metric}
  \centering
  \resizebox{1\linewidth}{!}{
  \begin{tabular}{ccc|ll}
    \toprule
    Arm & Amp Aug & Object & PA-MPJPE $\downarrow$ & MPJPE $\downarrow$ \\
    \midrule
    \newcheckmark & \newcheckmark & \newcheckmark & 1.02 & 1.99  \\
     & \newcheckmark  &  \newcheckmark & 1.07 \@{\hskip 0.06in}\textbf{-0.05 (-5\%)} & 2.22 \@{\hskip 0.06in}\textbf{-0.23 (-12\%)} \\
    \newcheckmark &  & \newcheckmark & 1.11 \@{\hskip 0.06in}\textbf{-0.09 (-9\%)} &  2.42  \@{\hskip 0.06in}\textbf{-0.43 (-22\%)} \\
    \newcheckmark & \newcheckmark &  & 1.07 \@{\hskip 0.06in}\textbf{-0.05 (-5\%)} & 2.13  \@{\hskip 0.06in}\textbf{-0.14 (-7\%)}  \\
    \halfcheckmark & \newcheckmark & \halfcheckmark & 1.04 \@{\hskip 0.06in}\textbf{-0.02 (-2\%)} & 2.07 \@{\hskip 0.08in}\textbf{-0.08 (-4\%)} \\
   \bottomrule
   \end{tabular}
  }
\end{table}

\section{Comparison with Data from Generative Models}\label{J}

\begin{figure}[t]
  \centering
  \includegraphics[width=1\linewidth]{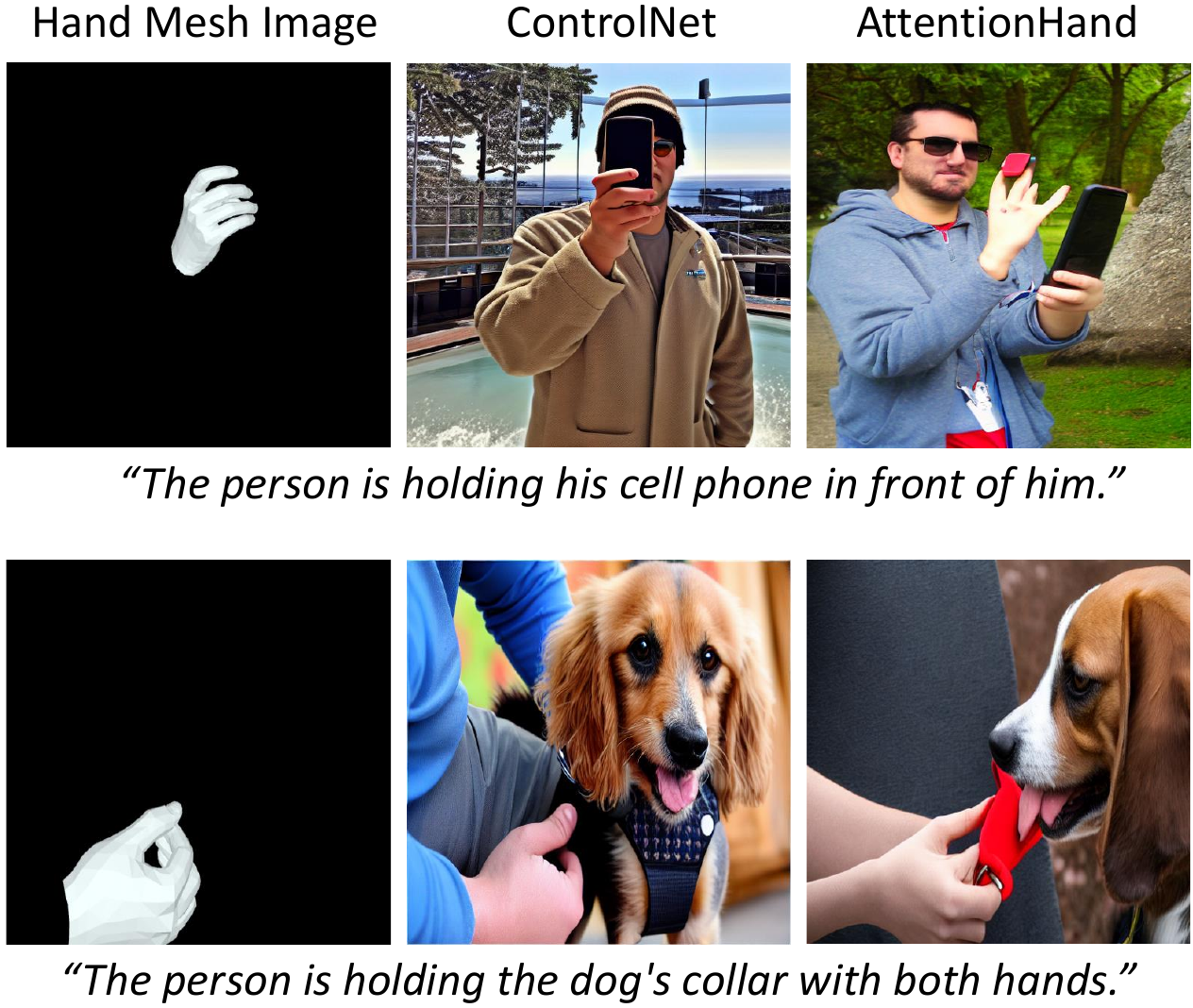}
  \caption{Synthetic hand data generated by text-to-image generative models with the hand mesh image control and text prompt.}
  \label{fig:comparison_gen}
\end{figure}

We run the official code of ControlNet~\cite{zhang2023adding} and AttentionHand~\cite{park2024attentionhand} to generate synthetic hand data given mesh images and text prompts. Since the pre-trained checkpoint is not open-sourced by AttentionHand currently, we train it from scratch. We also train the ControlNet given hand mesh images and text prompts as conditions. MSCOCO~\cite{lin2014microsoft} is used as the training set following AttentionHand. As shown in Fig.~\ref{fig:comparison_gen}, although the generated hand images have realistic backgrounds, the hands do not align well to the given mesh prompts. While generative models have significant potential for data synthesis, further research is needed to explore how to produce large-scale, plausible, and reliable synthetic data effectively.

\section{Additional Visualizations}\label{K}
\begin{figure*}[tb]
  \centering
  \includegraphics[width=1\textwidth]{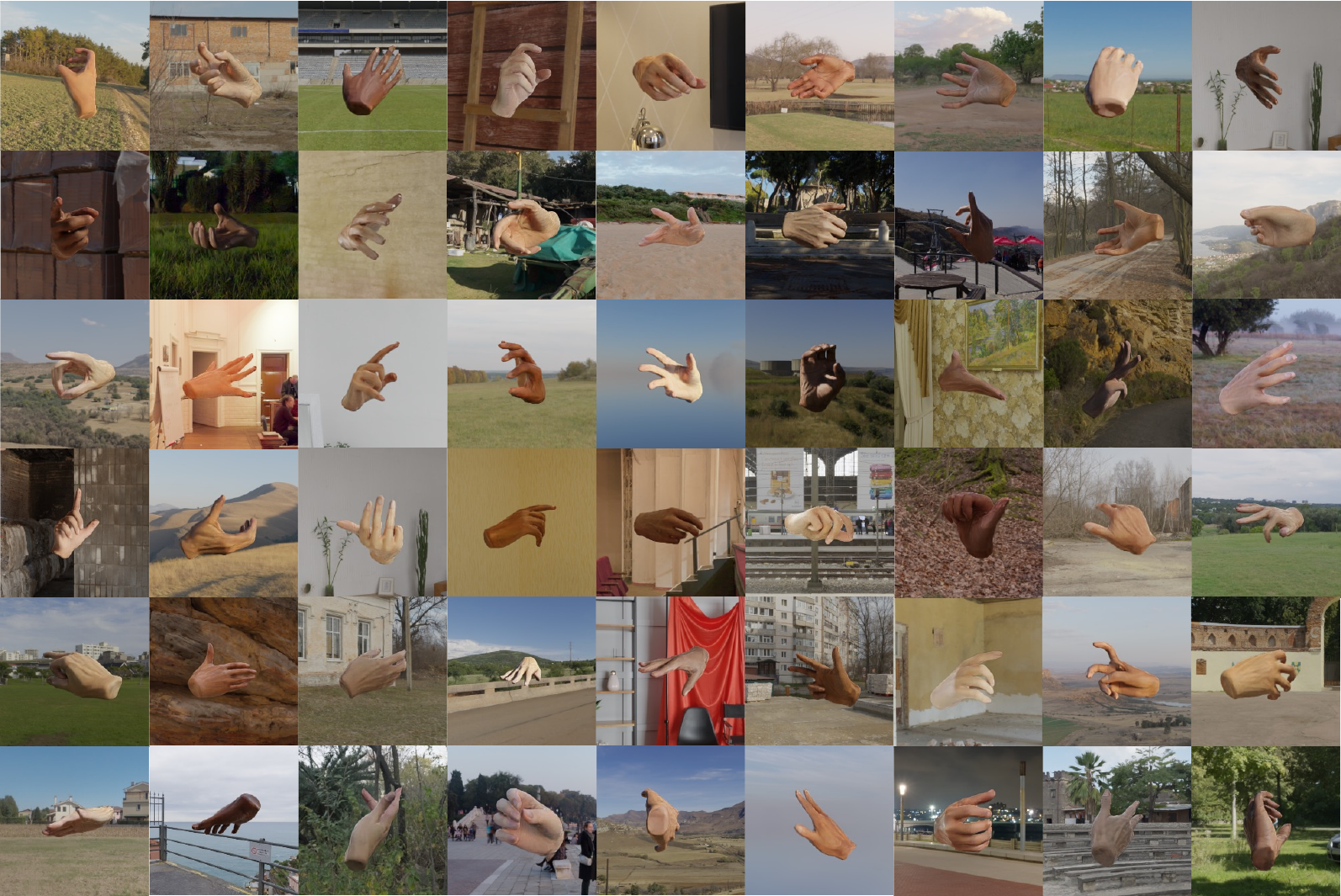}
  \caption{Synthetic hand image samples in our dataset.}
  \label{fig:dataset1}
\end{figure*}

\begin{figure*}[h]
  \centering
  \includegraphics[width=1\textwidth]{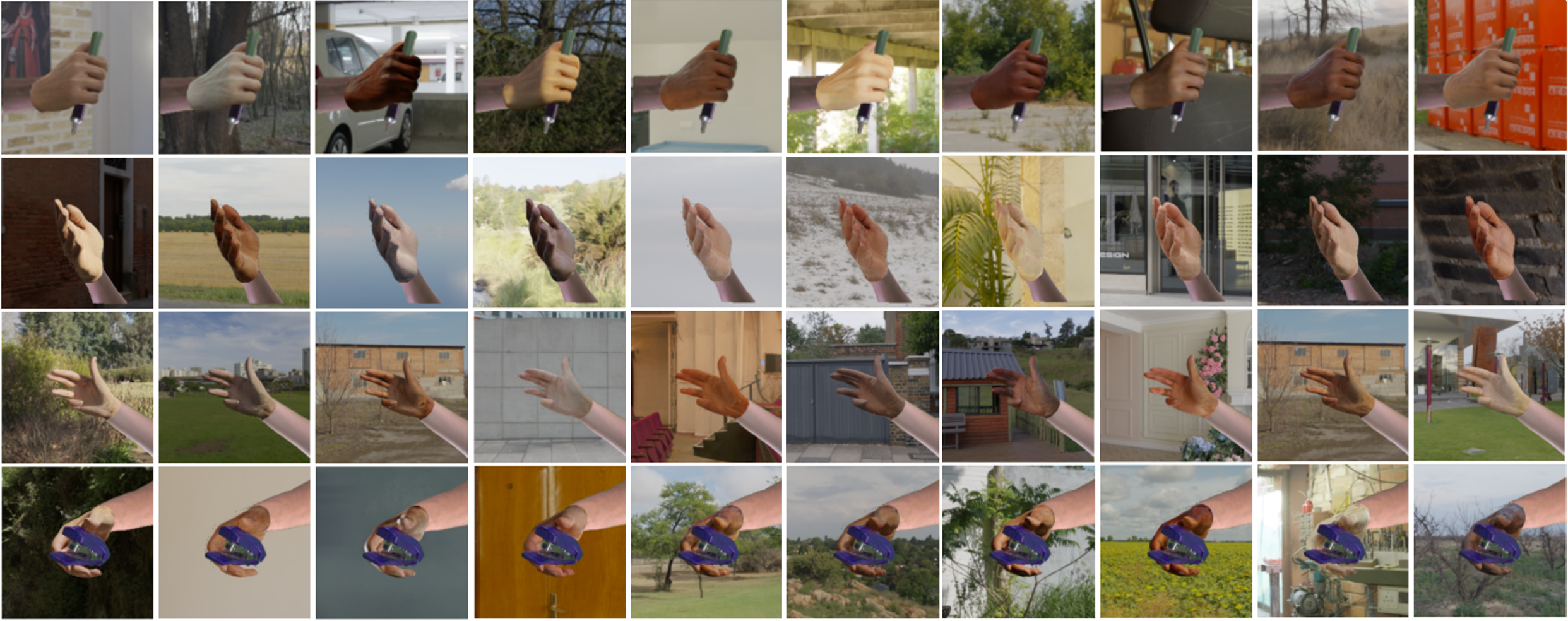}
  \caption{Synthetic hand images with arms or interacting objects.}
  \label{fig:dataset2}
\end{figure*}

Fig.~\ref{fig:dataset1} and Fig.~\ref{fig:dataset2} show some examples of synthetic hand images and synthetic hand images with arms and interacting objects. 

\end{document}